\documentclass[10pt,twocolumn,letterpaper]{article}

\usepackage{cvpr}
\usepackage{times}
\usepackage{epsfig}
\usepackage{graphicx}
\usepackage{amsmath}
\usepackage{amssymb}

\usepackage{algorithm}
\usepackage[noend]{algpseudocode}
\usepackage{caption}

\makeatletter
\def\BState{\State\hskip-\ALG@thistlm}
\makeatother

\usepackage[breaklinks=true,bookmarks=false]{hyperref}

\cvprfinalcopy 

\def\cvprPaperID{2946} 

\begin{document}
	
	\title{AMNet: Memorability Estimation with Attention}
	
	\author{Jiri Fajtl$^1$, Vasileios Argyriou$^1$, Dorothy Monekosso$^2$, Paolo Remagnino$^1$\\ 
		$^1$Kingston University, London, UK\\
		$^2$Leeds Beckett University, Leeds, UK\\
	}
	
	\maketitle
	
	\begin{abstract} 
		In this paper we present the design and evaluation of an end-to-end trainable, deep neural network  with a visual attention mechanism for memorability estimation in still images. We analyze the suitability of transfer learning of deep models from image classification to the memorability task. Further on we study the impact of the attention mechanism on the memorability estimation and evaluate our network on the SUN Memorability and the LaMem datasets. Our network outperforms the existing state of the art models on both datasets in terms of the Spearman's rank correlation as well as the mean squared error, closely matching human consistency.
		
	\end{abstract}

	\section{Introduction}
	The ability of man cognition to recall as well as forget visual content after viewing it is very important to the way we acquire new information and interact with our environment. This is becoming increasingly significant as creating and consuming visual content dominates other forms of information exchange. 
	Moreover, low cost, automated image and video capture systems are rapidly surfacing as the norm in the Internet of the Things (IoT) domain, also contributing to the visual information flow. 
	
	To which degree an image is later remembered or forgotten is expressed as image memorability. It is an important cognitive measure to be taken into account while processing visual content, whether for human to human or machine to human communication or for storage. 
	
	Memorability estimation has a large variety of practical applications, such as selecting or designing highly memorable advertising material, organizing and tagging of photos in albums, introducing a real-time, image memorability measure built into consumer digital cameras, helping to make highly memorable presentations and data visualizations, improving memorability of specific parts of a graphical user interface (GUI) or helping to illustrate education material. 
	An application of a great interest is to measure a decline in memory capacity of patients affected by dementia (such as Alzheimer's and Parkinson's diseases) and forms of mild cognitive impairment (MCI). 
	
	\begin{figure}[t]
		\begin{center}
			\includegraphics[width=1.0\linewidth]{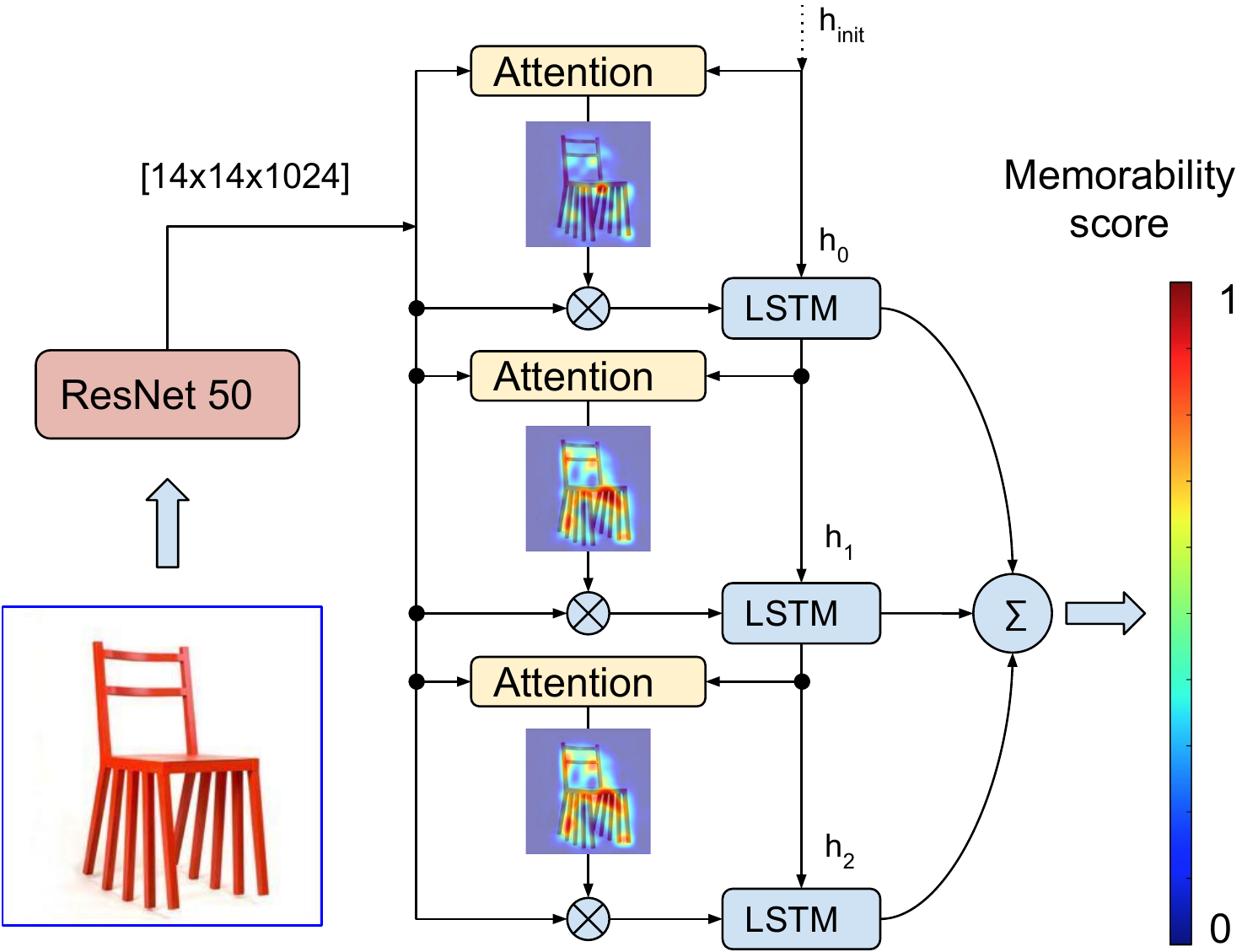}
		\end{center}
		\caption{AMNet iteratively generates attention maps linked to the image regions correlated with the memorability. After three iterations the memorability scores are added and presented on the output.}
		\label{fig:amnet}
	\end{figure}

Prior research \cite{what_makes_image_memorable_isola2011} has shown that image memorability has a stable property, that is, individuals tend to remember the same images with the same probability regardless of delays, and that it can be  quantified and measured. This research has led to first attempts to learn and predict memorability with machine learning freameworks, initially with low-level, global image features \cite{what_makes_photos_memorable}, reaching moderate success.	
To improve such a solution would, however, require the design of new features, which demands a strong domain knowledge not well understood in the specific case of memorability.

In \cite{memorability_with_deep_features}, \cite{memonet} and recently \cite{memorability_through_adaptive_transfer} has been shown that this problem can be mitigated by applying deep learning techniques to the memorability domain. Deep learning, however, requires large training dataset which was not available until A. Khosla et al. \cite{khosla2015understanding} introduced a large memorability dataset LaMem with 60K images and subsequently used it to train the MemNet, which is based on the AlexNet \cite{alexnet} initialized on the ImageNet \cite{imagenet} and Places \cite{places_dataset} datasets. MemNet achieves Spearman's rank correlation $\rho=0.64$ compared with the human consistency $\rho=0.68$ as measured by \cite{khosla2015understanding}.

Intuitively, image regions immediately drawing our attention would appear to be linked with highly memorable visual content. Indeed, this assumption was confirmed  to be correct in the works of \cite{mancas2013memorability}, \cite{memorability_image_regions} and \cite{what_makes_image_memorable_isola2011} who already very early indicated a potential relationship between the visual attention and memorability but did not further investigate their correlation.
To that end, we propose the Attention based Memorability estimation Network-AMNet, a novel, deep neural network architecture with a recurrent, visual attention mechanism with the primary goal to improve on the state of the art for the memorability prediction task. We also show advantages of the visualization of the generated attention maps and their connection to the memorability property. 
Our approach is extensively evaluated on the LaMem \cite{khosla2015understanding} and SUN Memorability \cite{what_makes_image_memorable_isola2011} datasets. 
The main contributions of our work are:
	\begin{itemize}
		\item AMNet as a generic architecture for regression tasks with deep CNN, visual attention mechanism and recurrent neural network.

		\item application of the proposed AMNet to the image memorability estimation.
				
		\item introduction of the incremental memorability estimation with the recurrent network and demonstration of the achieved performance gain.

		\item introduction of the visual attention technique for the memorability estimation and presentation of the performance gain.

		\item demonstration that transfer learning from deep models, trained for image classification, is particularly beneficial for the memorability estimation.

	\end{itemize}
	
The paper is organized as follows:
Section 2 provides background material on image memorability, its properties, measurement and prediction. 
In section 3 we propose the AMNet 
and discuss the theoretical framework behind this architecture 
and the training procedure.
The performance of AMNet is studied in the section 4, with section 5 concluding this work.

\section{Previous Work}

In a pioneering work on image memorability, Isola et al. \cite{what_makes_image_memorable_isola2011}, \cite{isola2011understanding} demonstrated that the ability of our cognition system to remember certain images and forget other is congruent among independent observers, despite large variability in the image content, concluding memorability is a stable property, intrinsic to images. Based on this premise, Isola et al.  \cite{what_makes_image_memorable_isola2011} investigated factors that give rise  to the image memorability effect, which was then used to predict image memorability scores with a machine learning program, based on global image features GIST \cite{oliva2001modeling}, SIFT \cite {lowe2004distinctive}, HOG \cite {hog}, SSIM  \cite{shechtman2007matching} and pixel histogram. 

In order to build better computational models to learn and predict memorability, researchers analyzed the relationship between memorability and various visual factors  \cite{memorability_image_regions}, image classes \cite{what_makes_photos_memorable} and saliency \cite{what_makes_object_memorable}.
Bylinskii et al. \cite{bylinskii2015intrinsic} conducted a number of experiments to better understand the intrinsic and extrinsic effects on image memorability, concluding that the primary substrate of memorability lies in the intrinsic properties of images and all extrinsic effects contribute only marginally. 

Deep learning was first applied to the memorability problem by Baveye et al. \cite{memonet} who proposed a MemoNet model based on GoogLeNet \cite{GoogLeNet}   trained on the ImageNet \cite{imagenet} dataset. 
\cite{memorability_with_deep_features} used CNN features with SVR \cite{svr} to predict  memorability with accuracy comparable to MemoNet \cite{memonet}.

To achieve higher accuracy with deep learning techniques Khosla et al. \cite{khosla2015understanding} collected a large memorability dataset LaMem with 60K images and introduced MemNet model based on the Hybrid-CNN, which is the AlexNet \cite{alexnet} CNN pretrained on the ImageNet \cite{imagenet} and the Places \cite{places_dataset} datasets (${\sim}3.6$ million images in total).  
Researchers also tried to improve memorability prediction by other techniques, such as the adaptive transfer learning from external sources \cite{memorability_through_adaptive_transfer} or predicting image memorability by multi-view adaptive regression \cite{peng2015predicting}, none exceeding the performance of the MemNet \cite{khosla2015understanding}. 

Relationship between the visual attention and memorability was already suggested by Isola et al. \cite{what_makes_image_memorable_isola2011} but was not further investigated. 
Mancas and Le Meur \cite{mancas2013memorability} studied the link between saliency and memorability and found that the most memorable images have  uniquely localized regions, while less memorable either do not have precise regions of interest or have several of them. Based on these findings, 
\cite{mancas2013memorability} devised new attention-related features that improved the memorability prediction by 2\% compared to the non attention based models from \cite{what_makes_image_memorable_isola2011}. 
In a similar work, 
Celikkale et al. \cite{celikkale2013visualattention} applied an attention driven spatial pooling pipeline based on SIFT \cite{lowe2004distinctive} and HOG \cite{hog} features and  bottom-up and object-level saliency detectors. Their results, albeit only moderate, still indicate a benefit of the attention based approach.
Importance of the memorability regions was explored by Khosla et al. \cite{memorability_image_regions} who introduced the concept of attention maps that relate image regions to memorability. These maps are learnt directly as clusters of gradients, textures and color features with the SVM-Rank solver \cite{joachims2006training} with results showing benefits of the attention on  memorability prediction.

In our work we investigate the application of deep learning methods with visual attention and recurrent network to learn and predict image memorability. To our knowledge the presented approach has not been attempted before.

\section{Method}
The idea behind the AMNet architecture is based on four main components a deep CNN trained on large-scale image classification task, a soft attention network, a Long Short Term Memory (LSTM) \cite{lstm_space_odyssey} recurrent neural network followed by a fully connected neural network for memorability score regression.

\begin{figure*}
	\begin{center}
		\includegraphics[width=.9\linewidth]{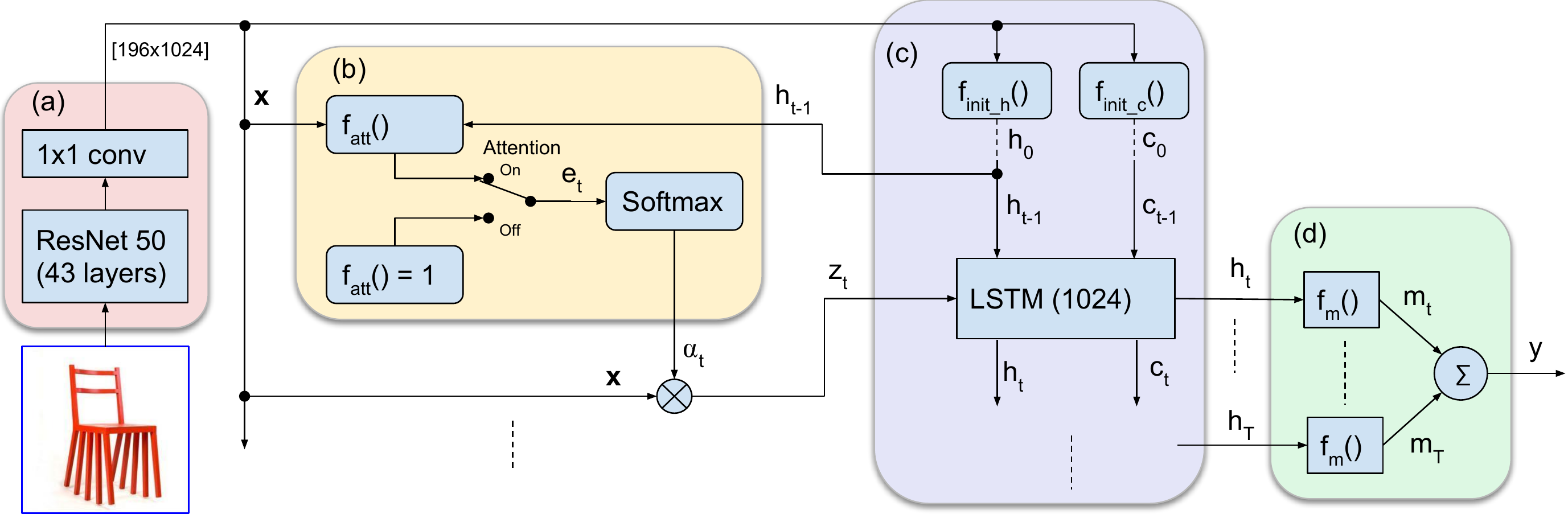}
	\end{center}
	\caption{A pretrained RestNet50 $(a)$ is followed by the soft attention mechanism $(b)$ with LSTM $(c)$, which over a sequence of three steps $T=3$ produces attention maps, each conditioned on the previous LSTM state $\mathbf{h}_{t-1}$ and the entire image feature vector $\mathbf{x}$. Memorability $y$ is then calculated as a sum of discrete memorability scores in the regression network $(d)$.}
	\label{fig:amnet-detail}
\end{figure*}

In the following section we introduce the details of the AMNet architecture as shown in Figure \ref{fig:amnet-detail}, starting with the pre-trained CNN $(a)$ 
for transfer learning.
Subsequently we show the working of the visual, soft attention mechanism $(b)$, the LSTM and network for the memorability regression $(c)$ and $(d)$. Finally we outline the training procedure and finish with the data augmentation process.

\subsection{Transfer Learning for Memorability Estimation}
It is common practice to use a pretrained CNN as a fixed feature extractor or to fine tune it for a similar application \cite{cnn_features_off_the_shelf}, mainly to reduce training time and overfitting on tasks with small datasets.

This technique is readily applied to computer vision problems centered around semantic features such as objects detection and segmentation, however little is known about such transfer learning for the image memorability estimation since there is no clear understanding of what visual features trigger the effects of remembering and forgetting. 

Khosla et al. \cite{khosla2015understanding} has already shown the benefits of fine tuning of pretrained CNN for this domain, however we decided to evaluate a much deeper model as a fixed feature extractor. Our results show that the features learnt for image classification are highly suitable for the memorability task. In our work we use ResNet50  \cite{residual_net} model trained on ImageNet where it achieves the top 1 error 24.7\%.

\subsection{Soft Attention Mechanism}
The ability of a neural network to learn which discrete information elements to focus on within a given training sample was first applied in machine translation by Bahdanau et al. \cite{bahdanau2014_neural_machine_translation}. This mechanism is called soft attention due to the fact that it produces a  probability weight for every information element rather than a hard decision boundary. The benefit of soft attention is that it can be learnt end-to-end 

with a gradient based optimization method.

The soft attention mechanism has two components, a network that learns probabilities for each information element within the input data and a gating function that uses these probabilities to weigh data for further processing.

\subsection{AMNet Details}

The AMNet estimates the image memorability by taking a single image $\boldsymbol{X}$ and generating a memorability score $y$.
\begin{equation} \label{eq:mem_est_fcn}
y=f(\boldsymbol{X}), \quad   y=[0,1]
\end{equation}
The process of memorability estimation is summarized in algorithm \ref{alg:amnet}.

\begin{algorithm}
	\caption{AMNet algorithm}\label{euclid}
	\begin{algorithmic}[1]
		\Procedure{MEMORABILITY(X)}{} \Comment{$y=f(\boldsymbol{X})$}
		
		\State $\mathbf{x}=\textit{get\_cnn\_features}(\boldsymbol{X})$ \Comment{ResNet50 fwd pass}
		
		\State $\mathbf{h}_{0}=\textit{f}_{init_c}(\mathbf{x})$	\Comment{Eq. \ref{eq:lstm_init}}
		\State $\mathbf{c}_{0}=\textit{f}_{init_h}(\mathbf{x})$	\Comment{Eq. \ref{eq:lstm_init}}

		\State $lstm\_init(\mathbf{h}_0, \mathbf{c}_0)$
		\State $y=0$

		\For{$t = 0$ to ${T}$}	
											\Comment{at $t=0 \rightarrow \mathbf{h}_t=\mathbf{h}_0$}
		\State $\mathbf{e} = f_{att}(\mathbf{x}, \mathbf{h}_{t})$   \Comment{Eq. \ref{eq:fatt}}
		\State $\boldsymbol{\alpha} = softmax(\mathbf{e})$	\Comment{Eq. \ref{eq:softmax}}
		
		\State $ \mathbf{z} = []$
		\For{$i = 0$ to ${L}$} \Comment{for all locations, Eq. \ref{eq:z}} 
			\State $ \mathbf{z} = \mathbf{z} + {\alpha}_i \mathbf{x}_i$  \Comment{$\mathbf{z} \in \mathbb{R}^D$}
		\EndFor
		 				
		\State $\mathbf{h}_t, \mathbf{c}_t = lstm\_step(\mathbf{z}, \mathbf{h}_t, \mathbf{c}_t)$ \Comment{Eq. \ref{eq:lstm}}
		
		\State $y = y + f_m(\mathbf{h}_t)$ \Comment{Eq. \ref{eq:memity_out}}
		\EndFor
		
		\State \textbf{return} $y$ \Comment{Memorability score $[0,1]$}
		
		\EndProcedure
	\end{algorithmic}
	\label{alg:amnet}			
\end{algorithm}

Formally, let the image features, extracted by a CNN, be a tensor with dimensions $(W,H,D)$ where $W$ and $H$ represent the spatial resolution while $D$ a length of feature vectors, one for each location within the $(W,H)$ region. Specifically, in the case of AMNet the feature tensor has dimensions $14\times14\times 1024$. In general there are $L=W\times H$ locations, represented as a vector $\mathbf{x}$:
\begin{equation} \label{eq:1}
\mathbf{x} = \left\lbrace \mathbf{x}_1,...,\mathbf{x}_L\right\rbrace \qquad  \mathbf{x}_i \in \mathbb R^{D }
\end{equation}
All vectors are column vectors, unless stated otherwise. The memorability is estimated with LSTM \cite{lstm_space_odyssey} over a three steps long sequence $T=3$. The LSTM is defined as:
\begin{equation} \label{eq:lstm}
\mathbf{h}_t = \phi(\mathbf{h}_{t-1}, \mathbf{z}_t)  \qquad  t=[0,T), h  \in \mathbb R^{B  }
\end{equation}
where $\mathbf{h}_t$ is the LSTM state at time $t$ with size $B=1024$.
The vector $\mathbf{z}_t$ represents a new image features  produced at the step $t$ as a result of the application of the attention weights $\boldsymbol{\alpha}^t$ on the input image features $\mathbf{x}$ and is calculated as a simple weighted sum  such that

\begin{equation} \label{eq:z}
\mathbf{z}_t = \sum_{i=1}^{L} {\alpha}_{t,i} \mathbf{x}_i  \qquad \mathbf{z}_t \in \mathbb{R}^{D  }
\end{equation}
where $\boldsymbol{\alpha}$ are the attention probabilities conditioned on the entire image feature vector $\mathbf{x}$ and previous LSTM hidden state $\mathbf{h}_{t-1}$
\begin{equation} \label{eq:2}
\boldsymbol{\alpha}_t \sim p(\boldsymbol{\alpha}_t | \mathbf{x}, \mathbf{h}_{t-1}) \qquad \boldsymbol{\alpha}_t \in  \mathbb R^L
\end{equation}
The attention probabilities, as well as other functions are parameterised with neural networks. The attention is then represented as a vector of weights produced by a softmax function 
\begin{equation} \label{eq:softmax}
{\alpha}_{t,i}={\frac {\exp({{e}_{t,i}})}{\sum _{k=1}^{L}\exp({{e}_{t,k}})}} \qquad 
\end{equation}
The attention weights vector $\mathbf{e}_{t}$ is a product of the image feature vector $\mathbf{x}$ and the LSTM hidden state $\mathbf{h}_{t-1}$
\begin{equation} \label{eq:e}
{e}_{t,i} = f_{att}(\mathbf{x}_i, \mathbf{h}_{t-1}) 
\end{equation}
$f_{att}()$ is s simple sum of two affine transformations followed by logistic function
\begin{equation} \label{eq:fatt}
f_{att}(\mathbf{x}_i, \mathbf{h}_{t-1}) = \boldsymbol{M}_i\tanh(\boldsymbol{U}\mathbf{h}_{t-1} + \boldsymbol{K}\mathbf{x}_i + \mathbf{b})
\end{equation}
where $\boldsymbol{M}_{L\times D}, \boldsymbol{U}_{D\times B}, \boldsymbol{K}_{D\times D}$ and $\mathbf{b}_{D\times 1} $ are network weights and biases respectively,  estimated together with other parameters of the network during optimization.

In order to experiment with the effects of the attention we can conditionally disable it by defining the $f_{att}()$ as a constant function with unit output such that:
\begin{equation} \label{eq:fatt_unity}
f_{att}(\mathbf{x}_i, \mathbf{h}_{t-1}) = {1}
\end{equation}
The results it that all feature vectors in $\mathbf{x}$ 
are considered equally, thus disabling the attention mechanism.

At each step $t$ the network produces one discrete memorability score $m_t$ calculated as:
\begin{equation} \label{eq:memity_discrete}
m_t = f_m(\mathbf{h}_t)
\end{equation}
The function $f_m()$ maps the LSTM hidden state $\mathbf{h}_t$ to the memorability score $m_t=[0,1]
$. It is implemented as a two-layer neural network for regression with a single output neuron and linear activation function. Finally, the total image memorability score  $y$ is calculated as a sum of the discrete memorabilities $m_t$
\begin{equation} \label{eq:memity_out}
y = \sum_t^T m_t
\end{equation}
In the first step, the LSTM hidden $\mathbf{h}_0$ and memory $\mathbf{c}_0$ states are initialized from the image feature vector $\mathbf{x}$ as follows:
\begin{equation} \label{eq:lstm_init}
\mathbf{c}_0 = f_{init_c}\biggl(\frac{1}{L} \sum_{i}^{L}\mathbf{x}_i \biggr) \quad
\mathbf{h}_0 = f_{init_h}\biggl(\frac{1}{L} \sum_{i}^{L}\mathbf{x}_i\biggr)
\end{equation}
where the $f_{init}()$ functions are single, fully connected neural networks with $\tanh()$ activation.

\subsection{Training Procedure}
The AMNet model is trained by minimizing the following loss function:
\begin{equation} \label{eq:2}
\mathcal{L} =  ( \hat  {y}  - y  )^{2}  + \lambda \mathcal{L}_\alpha
\end{equation}
The first term represents a mean squared error between the ground truth $\hat y$ and predicted image memorability $y$.
In order to encourage the attention model to explore all image regions over all time steps, we add a second term $\lambda\mathcal{L}_{\boldsymbol{\alpha}}$ which performs a joint $\ell_1$-$\ell_2$ penalty as a function of activations of all attention maps in the LSTM sequence $T$, introduced by Xu et al. \cite{show_attend_tell}. The hyper-parameter $\lambda$ specifies the impact of this penalty.
\begin{equation} \label{eq:L_alpha}
\mathcal{L}_{s{\alpha}}  =  \sum_{i}^{L} {s}_{i}^2
\end{equation}
${s}_{i}$ represents the $\ell_1$ penalty, which enforces sparsity along the sequence dimension $T$. In other words, it encourages a strong activation for 
only  one of the attention maps at location $i$.
\begin{equation} \label{eq:2}
{s}_{i} =  1-\sum_{t}^{T} {\alpha}_{t,i} 
\end{equation}
Finally, the $\ell_2$ penalty in the form of $\sum_i {s}_{i}^2$ in Eq.  \ref{eq:L_alpha} further promotes an even distribution of activations over all locations. 
The value of the $\lambda$ parameter was experimentally determined as $10^{-4}$
for which the network achieved the highest performance.

The entire model if fully differentiable and trained end-to-end with the ADAM \cite{adam} optimizer with a fixed learning rate $10^{-3}$. The input image feature vector $\mathbf{x}$ is extracted from the $43^{rd}$ layer of the RestNet50 \cite{residual_net} with dimensions $[14\times 14 \times 1024]$. The ResNet50 is trained for image classification on the ImageNet dataset and its weights are not updated during the AMNet training. 

\begin{figure}[t]
	\begin{center}
		\includegraphics[width=1.\linewidth]{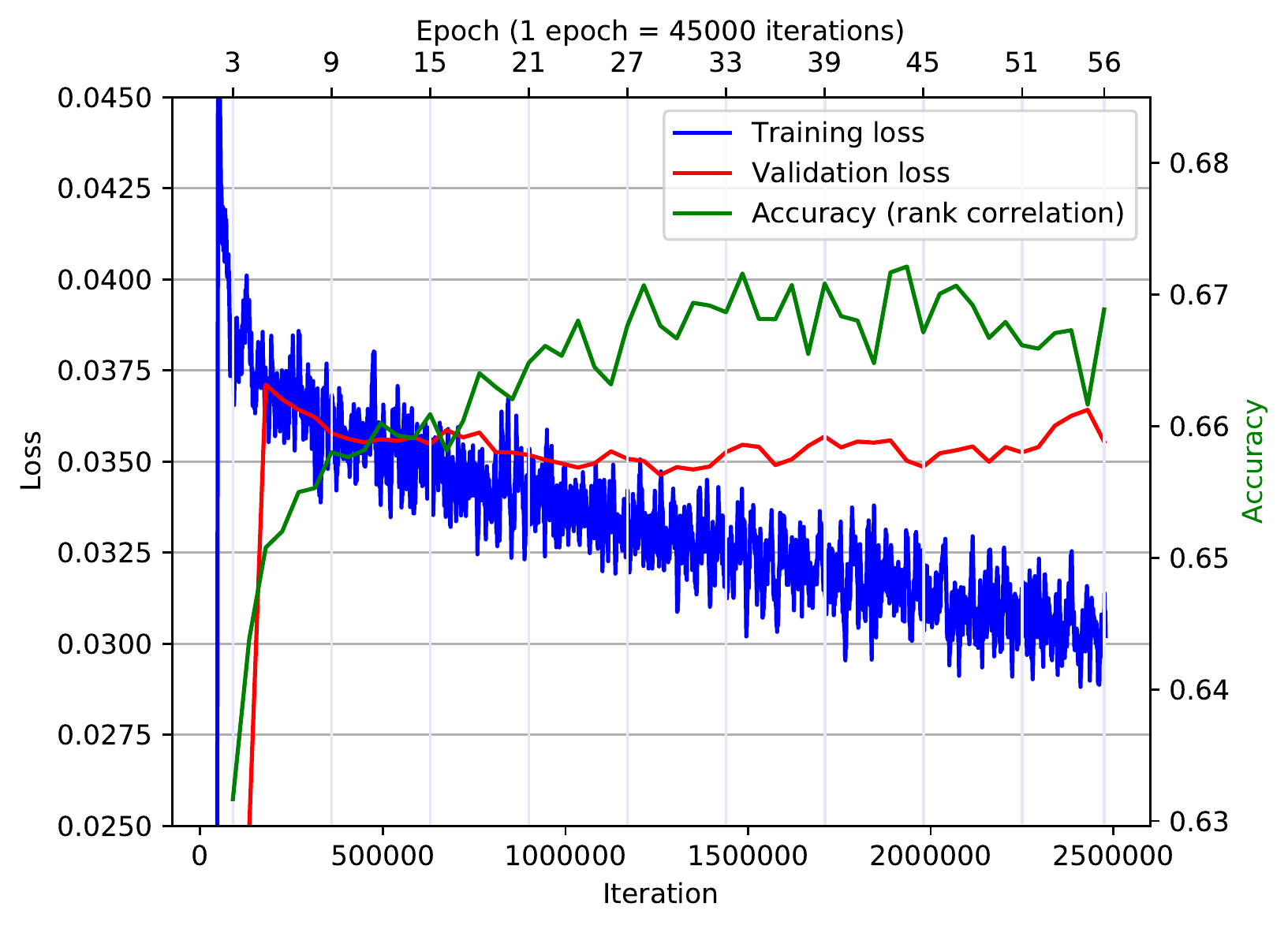}
	\end{center}
	\caption{Training/validation losses and memorability rank correlation on the validation dataset in the LaMem split1.}
	\label{fig:train_val_lamem_split1}
\end{figure}

The AMNet network is heavily regularized with dropout and with small $\ell_2$ weights regularization $10^{-6}$. We found that the dropout was critical to stop the network from overfitting. The training was carried out in minibatches of 256 images and terminated by early stopping when the observed Spearman's rank correlation on the validation dataset reached its maximum, which was 
between epoch 30 and 50 depending on the split and the training dataset (LaMem or SUN). Training and validation losses as well as the memorability rank correlation on the validation dataset in the LaMem, split 1 is shown in Figure \ref{fig:train_val_lamem_split1}.

\subsection{Data Preprocessing and Augmentation}
Common augmentation techniques are applied to the images during the training stage to reduce overfitting and improve generalization. 
A crop of random size of (0.08 to 1.0) of the original size and a random aspect ratio of 3/4 to 4/3 of the original aspect ratio is made and then resized to $224\times 224$ and randomly, horizontally flipped. 
For the evaluation only a center crop $224 \times 224$ was selected for the input.

\begin{figure}[t]
	\begin{center}
		\includegraphics[width=1.\linewidth]{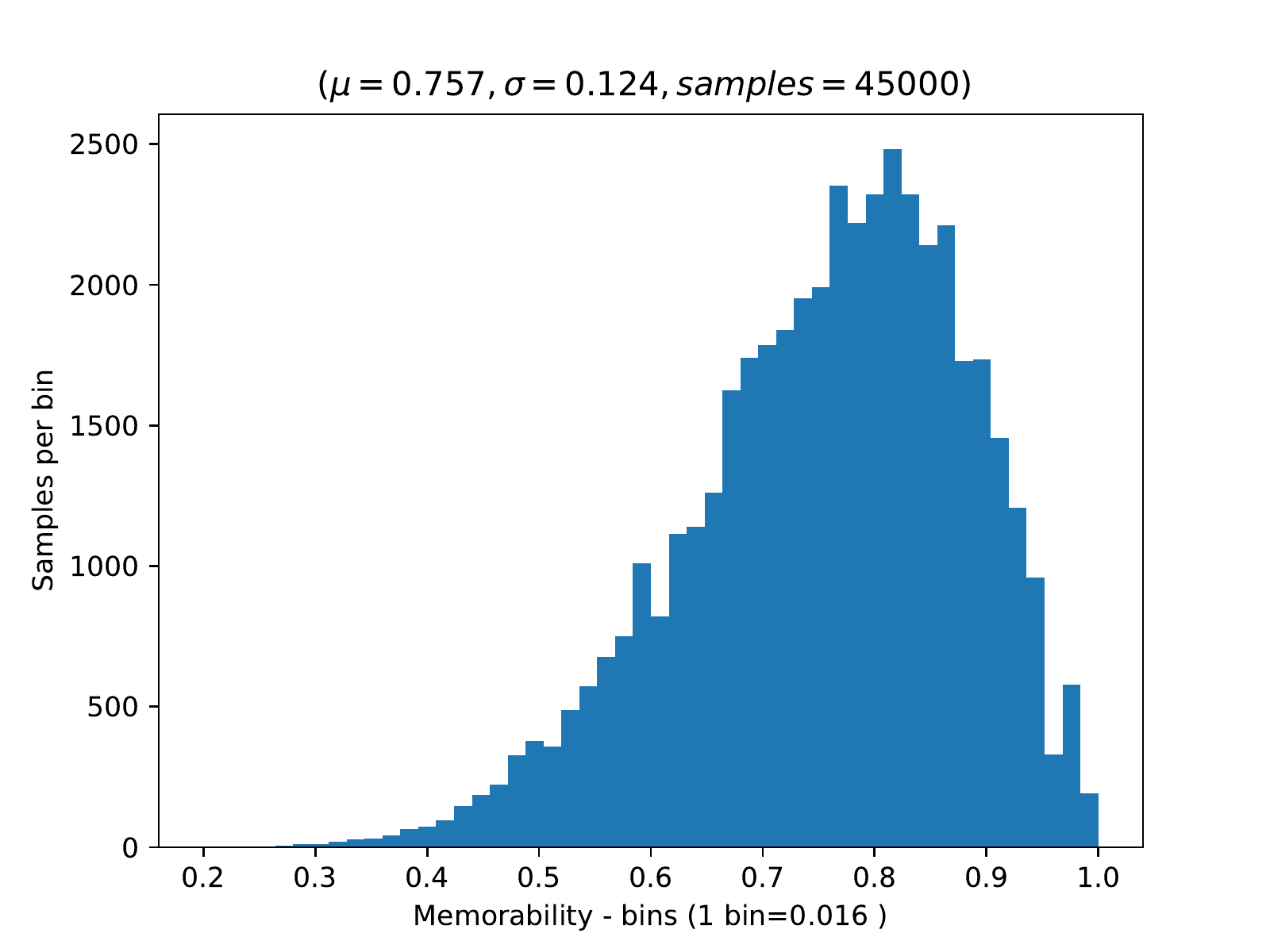}
	\end{center}
	\caption{Histogram of ground truth memorability scores in the LaMem \cite{khosla2015understanding} training dataset split1.}
	\label{fig:hist_lamem}
\end{figure}

Memorability scores in the LaMem dataset are in the range $[0,1]$ with distribution shown in Figure \ref{fig:hist_lamem}. For the training purpose the memorability scores were zero mean centered and scaled to range $[-1,1]$.

\section{Experimental Results}
In this sections we evaluate the AMNet on the LaMem \cite{khosla2015understanding} and SUN Memorability \cite{what_makes_image_memorable_isola2011} datasets. First we briefly describe the datasets and used evaluation metrics, and then present our qualitative and quantitative results with the comparison against the state of the art.

\subsection{Datasets}
\label{sec:datsets}
Main focus of this research work is on the LaMem \cite{khosla2015understanding} dataset due to its large size which makes it suitable for training deep neural networks.
The LaMem is the largest annotated image memorability dataset to this date with total of 58741 images. 
The images cover a wide range of indoor and outdoor environments, objects and people and were obtained from other labeled datasets such as MIR Flicker, AVA dataset \cite{ava_murray2012}, affective images dataset 
\cite{judd2009learning}, image saliency datasets 
\cite{machajdik2010affective}, 
\cite{ramanathan2010eye}, SUN 
\cite{xiao2010sun}, image popularity dataset \cite{khosla2014makes}, Abnormal Objects dataset \cite{saleh2013object} and a Pascal dataset \cite{farhadi2009describing}. The memorability scores were collected manually on the Amazon Mechanical Turk (AMT) by means of a memorability game introduced by \cite{what_makes_image_memorable_isola2011} and improved by \cite{khosla2015understanding}. Approximately 80 measurements (memorable=yes/no) were collected per image. There are 5 random splits each with 45000 images for training, 3741 for evaluation and 10000 for testing.

As a second dataset for evaluation we chose the SUN Memorability dataset pioneered by Isola et al. \cite{what_makes_image_memorable_isola2011}. There are 2222 images in total, originating from the SUN \cite{xiao2010sun} dataset with memorability scores collected similarly to the LaMem. There are 25 random splits with equal number of 1111 images for training and testing. 

\subsection{Evaluation Metrics}
Following the previous work, we report on the performance in terms of rank correlation, specifically a Spearman's rank correlation coefficient \cite{spearman}  
$\rho$ and mean squared error $\operatorname {MSE}$.

The Spearman's rank correlation coefficient measures consistency between the predicted and ground truth ranking, within the range $[-1, +1]$ where zero represents no correlation.
Higher $\rho$ values indicate better memorability prediction method:

\begin{equation} \label{eq:spearman}
\rho_{s}({\hat{r}}, r)={1-{\frac {6\sum_i^N ({\hat{r_i}}-r_i )^{2}}{N(N^{2}-1)}}}
\end{equation}
where $N$ is a number of samples, ${\hat{r_i}}$ is a rank of the $i^{th}$ ground truth memorability score, and $r_i$ the $i^{th}$ prediction. 

$\operatorname {MSE}$ is used as a secondary metric, not always presented in previous work. The Spearman's rank correlation shows a monotonic relationships between the reference and observations but does not reflect the absolute numerical errors between them, which is then presented by $\operatorname {MSE}$ according to: 
\begin{equation} \label{eq:mse}
\operatorname {MSE}({\hat{y}}, y)={\frac  {1}{N}}\sum _{{i=1}}^{N}({\hat  {y_{i}}}-y_{i})^{2}
\end{equation}
where $\hat  y_{i}$ is the ground truth memorability score, while $y_{i}$ the prediction and $N$ number of tested samples.

\subsection{Performance Evaluation}
In order to obtain results that are fully comparable with the previous work, we used the same training and evaluation protocol as in the \cite{khosla2015understanding} for the LaMem dataset and   \cite{what_makes_image_memorable_isola2011} for the SUN memorability dataset.

Evaluation on the LaMem dataset was performed by training one model on each of the five random splits as suggested by the authors \cite{khosla2015understanding} and then reporting the final memorability rank correlation and $\operatorname {MSE}$
, averaged over the  results from five corresponding test datasets. 

\begin{table}
	\begin{center}
		\begin{tabular}{|l|c|c|}
			\hline
			\textbf{Method} (LaMem dataset) &  $\rho \uparrow$ & $\operatorname {MSE} \downarrow$ \\
			\hline
			
			\hline
			
			AMNet  & \textbf{0.677} &  \textbf{ 0.0082} \\ \hline
			AMNet (no attention) &  0.663 & 0.0085  \\ \hline
			MemNet \cite{khosla2015understanding} & 0.64 & NA \\ \hline
	\begin{tabular}[c]{@{}l@{}}
 	CNN-MTLES \cite{memorability_through_adaptive_transfer}\\ 
 	(different train/test (50/50) split)
 	\end{tabular} &  0.5025 & NA \\ 
			\hline			
		\end{tabular}
	\end{center}
	\caption{Average Spearman's rank correlation $\rho$ and  $\operatorname{MSE}$ over 5 test splits of the LaMem dataset.}
	\label{tab:results_lamem}	
\end{table}

In Table \ref{tab:results_lamem} we show that the AMNet model with the active attention achieves $\rho=0.677$, or a 5.8\% improvement over the best known method MemNet \cite{khosla2015understanding}. Even without attention the AMNet outperforms prior work by 3.6\% which demonstrates that the pretrained, deep CNN with our recurrent and regression network layers still achieve high accuracy. 
The comparatively low performance of the CNN-MTLES \cite{memorability_through_adaptive_transfer} method can be attributed to the fact that this model uses various, specifically engineered visual features and features extracted from CNN networks trained on ImageNet \cite{imagenet} and Places \cite{places_dataset}. Thus it does not leverage the end-to-end deep learning. The CNN-MTLES, however, uses the LaMem dataset, which indicates that even a large dataset does not significantly improve the performance of models based on engineered visual features.

\begin{table}
	\begin{center}
		\begin{tabular}{|l|c|c|}
			\hline
			\textbf{Method} (SUN Memorability dataset)&  $\rho \uparrow$ &  $\operatorname {MSE} \downarrow $\\			
			\hline
			
			\hline
			
			Isola \cite{what_makes_image_memorable_isola2011} & 0.462 & 0.017   \\ \hline
			Mancas \& Le Meur \cite{mancas2013memorability}  & 0.479 &  NA \\ \hline

			AMNet & \textbf{0.649} & \textbf{0.011}  \\ \hline
			AMNet (no attention)& 0.62 & 0.012  \\ \hline
			MemNet \cite{khosla2015understanding} & 0.63 & NA \\ \hline
			MemoNet 30k \cite{memonet} & 0.636 &  0.012\\ \hline
			Hybrid-CNN+SVR \cite{memorability_with_deep_features} & 0.6202 & 0.013\\ 
			\hline
		\end{tabular}
	\end{center}
	\caption{Evaluation on the SUN Memorability dataset. 
		All models were trained and tested on the 25 train/val splits.}
	\label{tab:results_sun}
\end{table}

To train the deep AMNet model on the rather small SUN dataset we had to increase regularization to avoid overfitting. 
We found that in this specific case  $\ell_2=10^{-4}$ weights regularization performed better than a stronger dropout or the combination of both. Table \ref{tab:results_sun} shows that the AMNet with attention performs 2\% better than the current best model. By disabling the attention the performance declined to $\rho=0.62$, demonstrating the advantages of visual attention for this task.

We found that during training  $\operatorname{MSE}$ on the validation datasets follows a similar trend with the rank correlation $\rho$, however the $\rho$ peaks after the model starts overfitting as seen in Figure \ref{fig:train_val_lamem_split1}. 
It is conceivable to assume that the slightly higher variance at the maximum  $\rho$ improves generalization in terms of the predicted and ground truth monotonic relationships, even though  $\operatorname{MSE}$ starts increasing. 
For example, during the training on the LaMem split 1, as seen in Figure \ref{fig:train_val_lamem_split1}, we attained maximum $\rho=0.6721$ and $\operatorname{MSE}=0.00848$ while $\rho=0.6676$ for minimum $\operatorname{MSE}=0.00844$.

\begin{figure}[t]
	\begin{center}
		\includegraphics[width=1.1\linewidth]{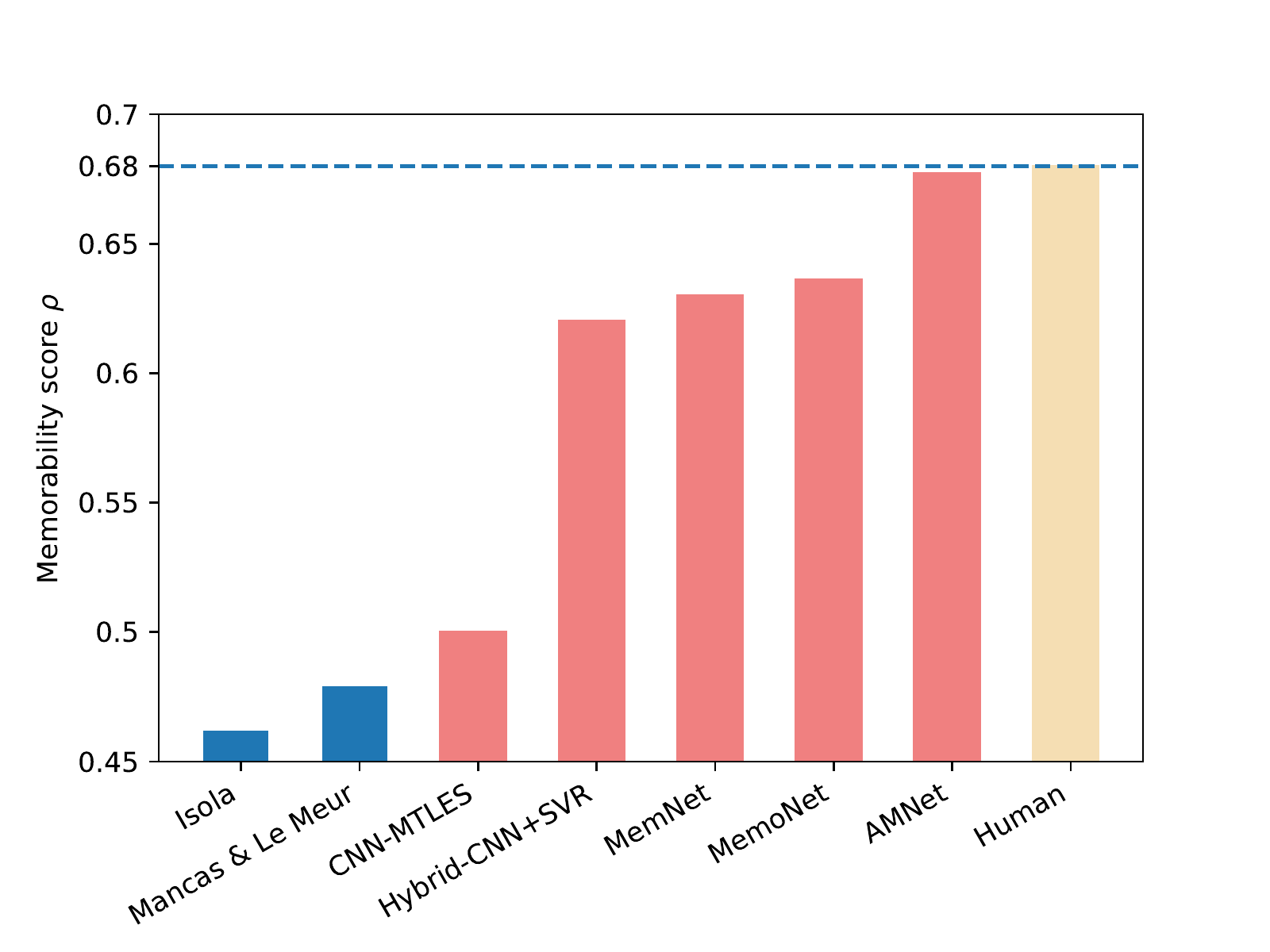}
	\end{center}
	\caption{Comparison against the state of the art methods. Red depicts deep learning based methods. AMNet, MemNet and CNN-MTLES \cite{memorability_through_adaptive_transfer} where trained on the LaMem, the rest on the SUN Memorability dataset.}
	\label{fig:comparison_sota}
\end{figure}

Tables \ref{tab:results_lamem} and \ref{tab:results_sun} show that the AMNet exhibits the best performance in terms of the Spearman's rank correlation as well as $\operatorname{MSE}$ on both, the LaMem and the SUN datasets. The best performance attains $\rho=0.677$ on the LaMem dataset, approaching $99.6\%$ of the human performance $\rho=0.68$ as measured by Khosla et al. \cite{khosla2015understanding}.
Comparison against the state of the art can be seeing in Figure \ref{fig:comparison_sota}.

\subsection{The Role of Attention on Memorability}
The significant performance gain is achieved by the fact that the neural network learns to focus its attention to specific regions most relevant to memorability. 
The improvement is close to 2\% on the LaMem and almost 5\% on the SUN dataset. 
AMNet learns to explore the image content by producing three visual attention maps, each conditioned on the image content obtained by  exploiting the previous map. We have experimented with 2,3,4,5 and 6 LSTM steps and found that three steps are sufficient to achieve the reported performance. 

\newlength{\himg}
\setlength{\himg}{2.28cm}

\begin{table*}[h!]
		
	\begin{center}
		\begin{tabular}{lllllll}
&
{\includegraphics[height=\himg]{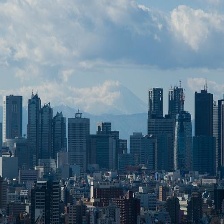}}& 
{\includegraphics[height=\himg]{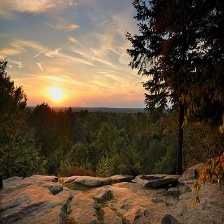}} & 
{\includegraphics[height=\himg]{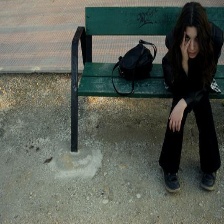}} & 
{\includegraphics[height=\himg]{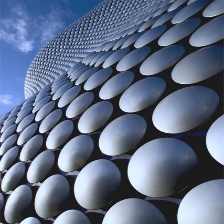}} & 
{\includegraphics[height=\himg]{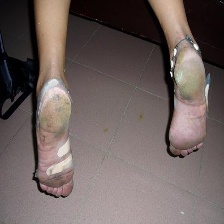}} & 
{\includegraphics[height=\himg]{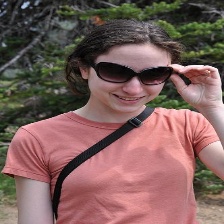}} \\ 
&a) 0.453 (0.45) & b) 0.453 (0.44) & c) 0.78 (0.794) & d) 0.881 (0.892) & e) 0.9 (0.894) & f) 0.887 (0.896) \\
\\		

t1 &
{\includegraphics[height=\himg]{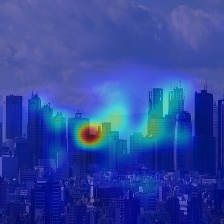}}&
{\includegraphics[height=\himg]{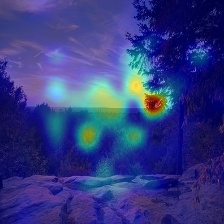}}&
{\includegraphics[height=\himg]{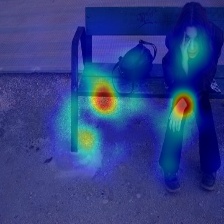}}&
{\includegraphics[height=\himg]{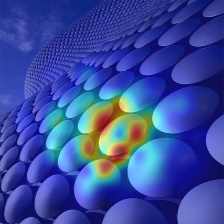}}&
{\includegraphics[height=\himg]{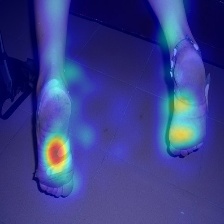}}&
{\includegraphics[height=\himg]{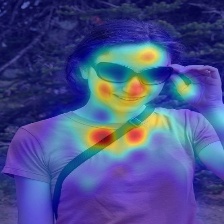}} \\
&0.165 & 0.167 & 0.256 &  0.289 & 0.293 & 0.290\\
\\
			 			
t2 &			
{\includegraphics[height=\himg]{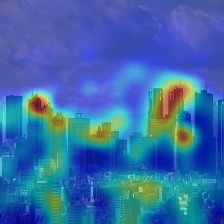}}&
{\includegraphics[height=\himg]{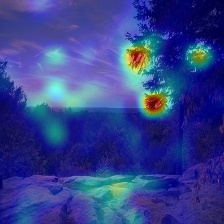}}&
{\includegraphics[height=\himg]{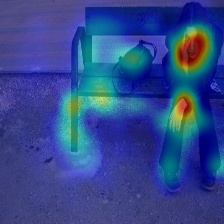}}&
{\includegraphics[height=\himg]{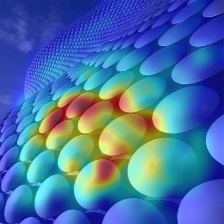}}&
{\includegraphics[height=\himg]{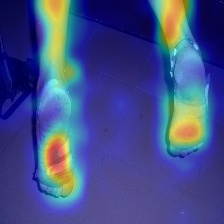}}&
{\includegraphics[height=\himg]{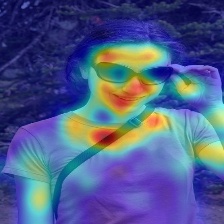}}\\
&0.148 &  0.148 & 0.262 &   0.295 & 0.302 & 0.297\\
\\
			
t3 &			
{\includegraphics[height=\himg]{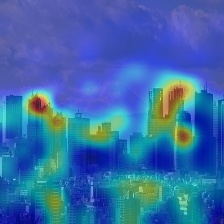}}&			
{\includegraphics[height=\himg]{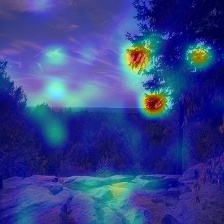}}&	
{\includegraphics[height=\himg]{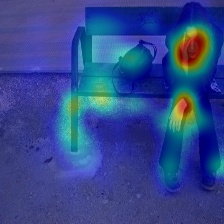}} &					
{\includegraphics[height=\himg]{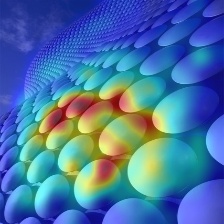}} &				
{\includegraphics[height=\himg]{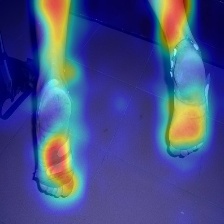}} &			
{\includegraphics[height=\himg]{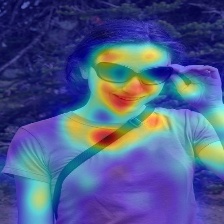}}\\
&0.140 & 0.128 & 0.262 & 0.298 & 0.306 & 0.300\\
			
			&
			{\includegraphics[height=\himg]{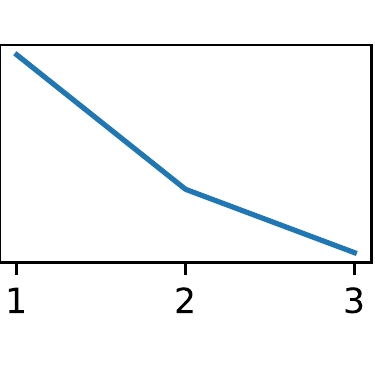}}&	{\includegraphics[height=\himg]{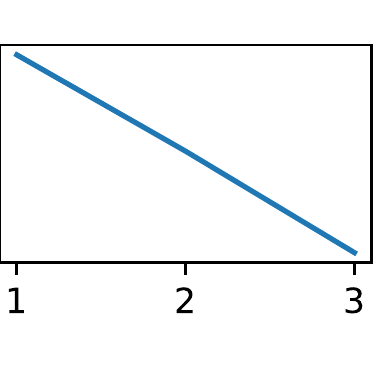}}&	{\includegraphics[height=\himg]{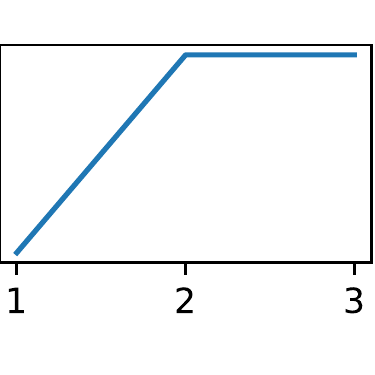}}&	{\includegraphics[height=\himg]{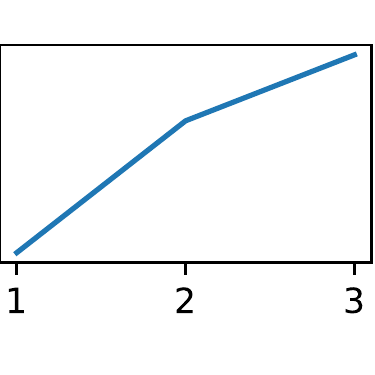}}&	{\includegraphics[height=\himg]{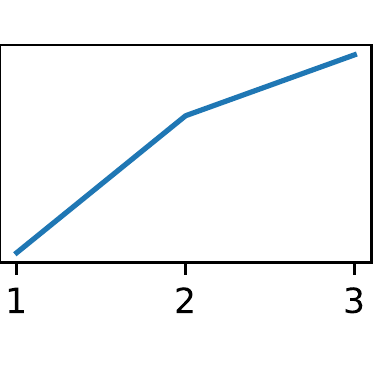}}&	{\includegraphics[height=\himg]{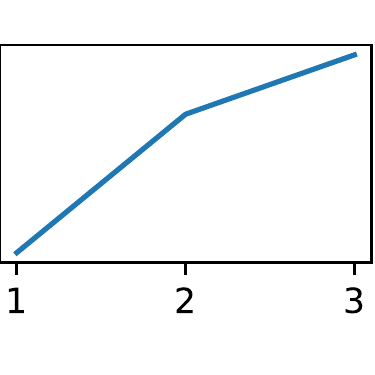}}			
		\end{tabular}
	\end{center}
	\captionof{figure}{Examples of attention maps for low and high memorability images from LaMem test dataset split 2. Tested images, their estimated and ground truth memorabilities (in brackets) are shown in the top raw. Bellow each image is a discrete memorability score estimated at the steps $t_1, t_2$ and $t_3$. Plots at the bottom row show gradients over three LSTM steps.}	
	\label{fig:att_maps}		
\end{table*}
In order to interpret the relation between the attention maps and corresponding discrete memorability estimations in each LSTM step, we converted the attention maps to heat maps and visualized them along with the memorability scores. In Figure \ref{fig:att_maps} we show selected images from the LaMem, split 2 test dataset. Images \emph{(a)}, \emph{(b)} and \emph{(c)} have low memorability, image \emph{(d)} a medium one and \emph{(e)} and \emph{(f)} high memorability. 
Images of the attention maps are obtained by taking  the output of the softmax function Eq. \ref{eq:softmax}, scaled to range $[0,255]$ and resized from $14\times 14$ to $244\times 244$.

As we can see in images \emph{(a)}, \emph{(c)} and \emph{(d)} in Figure \ref{fig:att_maps}, most of the first attention weights gravitate towards the image center, which is most likely caused by the Center Bias, studied in \cite{judd2009learning}, \cite{zhang2008sun} and attributed primarily to the photographer bias. In the subsequent LSTM steps, however, the attention usually moves to the regions responsible for  memorability. 

After a close inspection, we found that the attention maps for low memorability images tend to be sparser with few small peaks, while for higher image memorability, the attention maps display sharper focus covering larger regions around the activation peaks. 
Core image memorability usually originates in regions with people and human faces as evident in images \emph{(c)} and \emph{(f)} in Figure \ref{fig:att_maps}. 

\begin{figure}[t]
	\begin{center}
		\includegraphics[width=1.\linewidth]{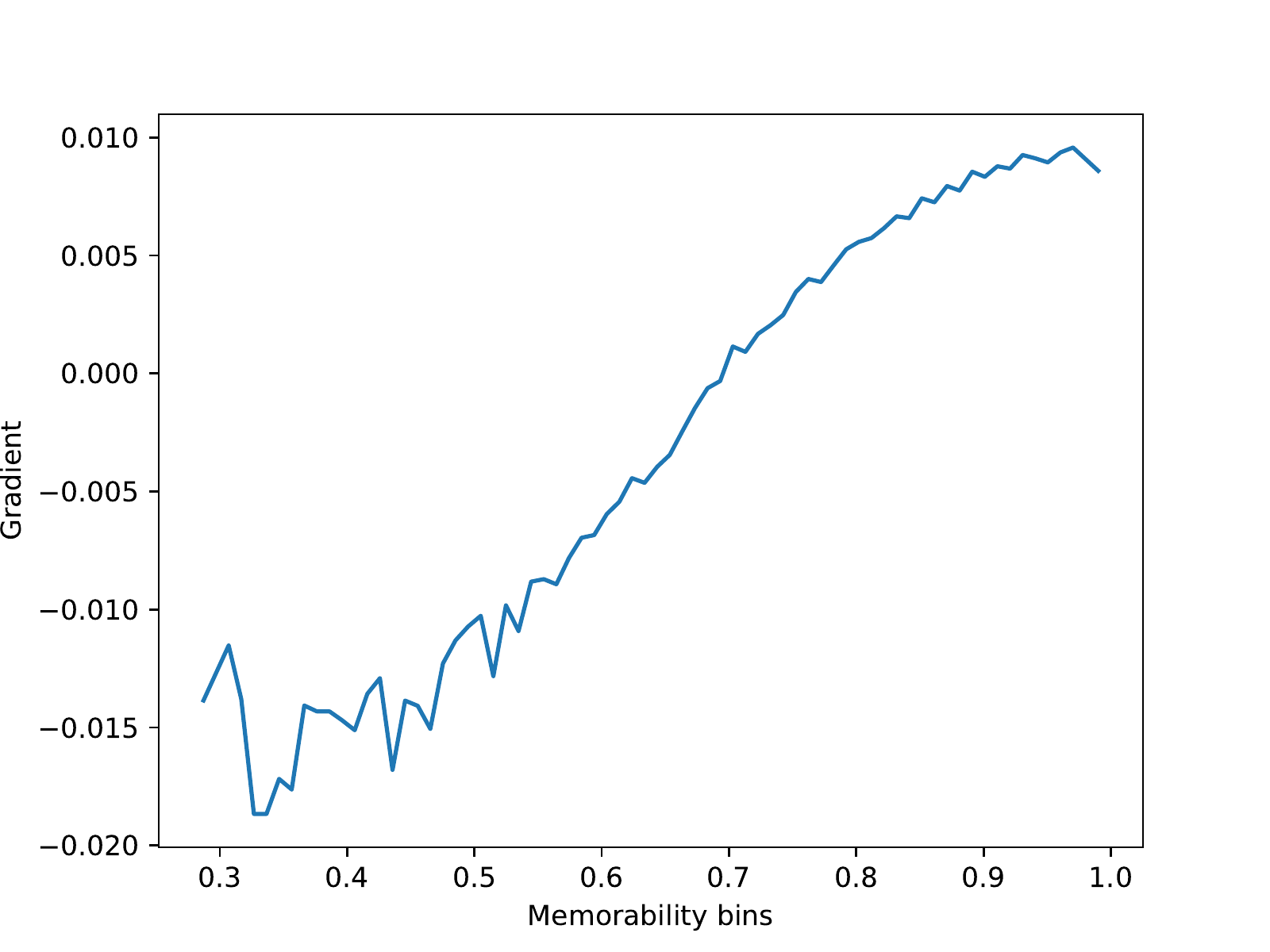}
	\end{center}
	\caption{Histogram of gradients of discrete memorabilities over the LSTM steps. The gradient is directly proportional to the total image memorability.}
	\label{fig:mem_gradient}
\end{figure}

Moreover, we found that the estimates of discrete memorabilities $m_t$ in Eq. \ref{eq:memity_discrete} decrease with each LSTM step $t$ for low memorability images, while for high memorability images they grow. This relation is shown in Figure \ref{fig:mem_gradient}. This effects is consistent within the LaMem test datasets across all splits and can be seen in Figure \ref{fig:att_maps}.

Initially, we experimented with additional penalty function that would encourage the optimizer to estimate the discrete memorabilities in ascending or descending order, however this always caused a drop in the performance. The above observation explains this effect, that is, the gradient of the discrete memorabilities over the LSTM steps differs depending on the core image memorability. Thus forcing the optimizer to maintain positive or negative gradient has a detrimental effect on the model convergence.


\section{Conclusion}
In this work we propose AMNet, a novel deep neural network with visual attention component for image memorability estimation. This network consists of a pre-trained, deep CNN followed by a modified visual attention mechanism with a recurrent network and network for memorability regression. By design the AMNet is generic and could be employed for other regression, computer vision tasks.

We show that a deep CNN, trained on large-scale image classification is beneficial for the memorability estimation task, indicating that the feature hierarchies extracted for the image classification are suitable to express the composition underlying the memorability effect.

Finally, we demonstrate that our recurrent visual attention network significantly improves performance of the image memorability learning and inference. 

The proposed method outperforms previous state of the art work by 5.8\% (from $\rho=0.64$ to $\rho=0.677$) on the Spearman's rank correlation and closely approaches the human performance $\rho=0.68$ with a 99.6\% consistency. The AMNet implementation in PyTorch is available at  \url{https://github.com/ok1zjf/amnet/}

{\small
	\bibliographystyle{ieee}
	\bibliography{ok1zjf}
}

\end{document}